\documentclass{ecai}
\usepackage{times}
\usepackage{latexsym}

\usepackage[latin1]{inputenc}
\usepackage{amssymb,amsmath,array}

\usepackage{tikz}
\usetikzlibrary{shapes,arrows,positioning}

\usepackage{algorithm}
\usepackage[noend]{algpseudocode}
\usepackage{bm}
\usepackage{paralist}
\usepackage{multirow}
\usepackage[inline]{enumitem}

\tikzstyle{invisible} = [circle, minimum width=0.5cm, minimum height=0.5cm, text centered]
\tikzstyle{layer}=[rectangle, rounded corners, minimum width=2cm, minimum height=0.1cm, text centered, draw=black]
\tikzstyle{red_layer}=[rectangle, rounded corners, minimum width=2cm, minimum height=0.1cm, text centered, draw=red]

\tikzstyle{concat} = [circle, minimum width=0.2cm, minimum height=0.2cm, text centered, draw=black]
\tikzstyle{arrow} = [thick,->,>=stealth]

\ecaisubmission   % inserts page numbers. Use only for submission of paper.
\begin{document}

\title{Pruning CNN's with linear filter ensembles}

\author{Csan\'ad S\'andor\institute{Babe\c{s}-Bolyai University, Cluj-Napoca, Romania, email: \{csanad.sandor, szabolcs.pavel, lehel.csato\}@cs.ubbcluj.ro} \textsuperscript{ ,}\institute{Robert Bosch SRL, Cluj-Napoca, Romania} \and Szabolcs P\'avel\footnotemark[1] \textsuperscript{ ,}\footnotemark[2] \and Lehel Csat\'o\footnotemark[1] }

\maketitle
\bibliographystyle{bibliography}

\begin{abstract}
  %We evaluate the pruning of neural networks in order to improve the quality and the execution speed of the system.
Despite the promising results of convolutional neural networks (CNNs), their application on devices with limited resources is still a big challenge; this is mainly due to the huge memory and computation requirements of the CNN.
To counter the limitation imposed by the network size, we use \emph{pruning} to reduce the network size and -- implicitly -- the number of floating point operations (FLOPs).
Contrary to the \emph{filter norm} method -- used in ``conventional'' network pruning -- based on the assumption that a smaller norm implies ``less importance'' to its associated component, we develop a novel filter importance norm that is based on the change in the empirical loss caused by the presence or removal of a component from the network architecture.

Since there are too many individual possibilities for filter configuration, we repeatedly sample from these architectural components and measure the system performance in the respective state of components being active or disabled.
The result is a collection of filter ensembles -- filter masks --  and associated performance values.
We rank the filters based on a linear and additive model and remove the least important ones such that the drop in network accuracy is minimal. 
We evaluate our method on a fully connected network, as well as on the ResNet architecture trained on the CIFAR-10 dataset.
Using our pruning method, we managed to remove $60\%$ of the parameters and $64\%$ of the FLOPs from the ResNet with an accuracy drop of less than $0.6\%$.

\end{abstract}

\section{Introduction}\label{sec:intro}

Modern neural networks process and extract information from large datasets. 
The information is stored in neural network weights; these weights -- given an architecture -- provide the outputs for the inputs.

A typical example is that of a deep -- modern -- network processing images and labeling them with positive or negative labels.
This network encodes the ``information'' in its weights and usually the number of weights in a deep network is of order of millions.
This high number of parameters is detrimental for at least two reasons:
\begin{enumerate}
\item usually, the architecture is too complex for the dataset: the number of network parameters is so large that it is likely that the data items are ``remembered'', without any \emph{generalization} taking place;
\item when executed, the calculation \emph{consumes} a lot of energy -- since it computes weighted sums with the huge parameter set.
\end{enumerate}
A proposal that -- at least partially -- relieves both problems is to \emph{eliminate} elements from the network architecture via pruning.
In pruning techniques a proposal to remove an architectural atom -- a single parameter from a neuron/filter, a neuron from a fully connected layer, a filter from a convolution layer -- is validated by first removing the respective atom, re-training the network, and observing whether the ``slimmer'' architecture is as good as the original -- un-pruned -- one.
The result of pruning is a system that provides (1) better generalization due to the reduced number of parameters and (2) reduced energy consumption also due to the smaller number of parameters.
%
%Despite the \emph{qualitative} promises that a pruning method makes, there are only a few \emph{principled} mains to attain it.

%the method is too expasive, not feasible
Unfortunately, examining importance of architectural atoms one by one is computationally unfeasible -- especially when the number of components is large.
Instead of this, most pruning techniques rank them based on their parameter norm (e.g. $l_2$ norm): the assumption is that components with small norm are less important so they can be removed from the network.
While this assumption assure fast importance evaluation, does not consider the network performance.
In this paper, we develop a novel filter importance norm that incorporates the network performance change caused by the elimination of multiple components from the CNN.
To measure the importance of a set of architectural components we associate an indicator to each. We consider a set of indicators and train our resulting model -- these indicators specify different filter ensembles -- measuring the performance in each case. The result is a collection of filter ensembles and \emph{associated} CNN performance.
Assuming a linear dependence between the filter ensembles, we rank the filters based on a linear and additive model and remove the least important ones such that the drop in network accuracy is minimal.

%In this paper we present a filter pruning technique that removes filters based on a novel filter importance estimation.

We have two main contributions in this work: (1) we introduce a new technique that is able to remove network components (neurons or filters) by considering the network performance; (2) we present our experiments where we compress networks trained on a synthetic and the CIFAR-10 dataset.

%%%%%%%%%%%%%%%%%%%%%%%%%%%%%%%%%%%%%%%%%%%%%%%%%%%%%%%%%%%
\section{Pruning methods}\label{sec:pruning}

We must first mention the pruning methods developed in the early days of neural networks.
One of the first attempts was termed ``Optimal Brain Damage''~\cite{Cun90optimalbrain} and ``Optimal Brain Surgery''~\cite{Hassibi:1993:OBS:2987189.2987223} where the Hessian of the loss function was used to simplify the network. 
The main disadvantage of these methods is the memory and computation cost due to the calculation of the Hessian matrix.
Since neural networks became popular again, different pruning methods were developed.
One of the main question is always how to find the redundant or less important parameters. 
Han et al. \cite{DBLP:conf/nips/HanPTD15} uses the magnitude of weights to perform intra-kernel pruning. 
It uses fine-tuning with $l_2$ regularization and iteratively removes parameters where magnitude is below a threshold. 
Louizos et al.~\cite{Louizos2017LearningSN} proposes a non continuous $l_0$ norm relaxation, that can be applied as a regularization term with stochastic gradient descent.
This way, some parameters of the network become zero during training and they can be removed afterwards.
The benefit of the above presented techniques is that the result is a highly compact network where the parameters are stored in sparse matrices.
The drawback is the need of special hardware for efficient inference due to the sparse representation~\cite{DBLP:journals/corr/HanMD15}.

To address the issue of sparse matrices, structured pruning were introduced. 
Here the main idea is to prune groups of parameters from the weight matrices, such as rows or columns.
For example, Li et al.~\cite{DBLP:journals/corr/LiKDSG16} prunes entire filters from a convolutional layer based on sensitivity analysis: it computes the $l_1$ norm of the filters (in a given layer), sorts them and sequentially removes the filters with small norm. 
At each filter removal it measures the accuracy of the network and stops the process when a significant accuracy drop is reached.
\cite{NIPS2016_6504} proposes a structured sparsity learning method using $l_1$ and $l_2$ norm to regularize filter, channel, filter shape and depth structures.
The method introduced by \cite{Luo_2017_ICCV} transforms the model pruning into an optimization problem and removes the channels by minimizing the next layer reconstruction error.
\cite{ijcai2018-309} applies soft pruning on the filters of the network.
The method ranks the filters in a given layer based on their $l_2$ norms and set the least important filters to zero (all the parameters of those filters).
Next they retrain the network such that the above filter values are updated too.
These two steps are repeated until some of the filter values converge to zero.
Finally, they remove the filters whose norm is close to zero.
He et al.~\cite{he2019filter} presents a new filter pruning criteria which uses the geometric median of the filters.
Instead of removing filters with small norm, this method calculates the geometric median of all the filters in a given layer and ranks them based on their distance from this point.
The authors say, that filters close to the median have similar contributions with the remaining filters thus the pruning does not affect the network accuracy.
A slightly different approach is presented by Zhang et al. in~\cite{10.1007/978-3-030-20867-7_24}, where structured pruning is applied on morphological neural networks.
In these networks the multiplication and addition operations are replaced by addition and maximum calculation (max-plus operator) which results in a non-linear computational unit.
Based on the parameters of the max-plus operator, the authors define a threshold for each max-plus unit and prune them based on the corresponding threshold.

Our work falls into the structured pruning category since we are eliminating entire neurons or filters -- thus no special hardware or library is required.
However, we estimate filter importance values by incorporating the network performance during the absence of different filters/neurons.

\section{The proposed method}\label{sec:pruning_probs}
%#motivation/problem definition
The aim of the network pruning process is to find and remove less important or redundant parameters from the network.
We consider a parameter less important and redundant if its absence does not result in a significant drop in validation accuracy or the original accuracy can be restored with network retraining.

Since we are interested in structured pruning, our goal is to remove entire filters or entire neurons.
For simplicity we mostly use the term filters and filter pruning but the same method can be applied in case of neurons as well.
The main question in network pruning is how to identify less important filters.
%
%brute force - remove some such that the accuracy drop is less than some prdefinec value
%not feasable
The most accurate method would be to measure the network accuracy with all possible filter combinations. 
Then remove filters such that the network size is sufficiently small and the accuracy is above a predefined threshold. 
Unfortunately, this method is not feasible since in case of $N_F$ filters this would require $2^{N_F}$ number of evaluations of the network.

%gready brute foce: remove each, measure accuracy drop and remove the one with which the accuracy drop is smallest
%still not feasable
%there could be cases when it is not the best: what if removing 2 filter causes less accuracy drop if we remove only one filter

A less computationally expensive heuristic is to temporarily remove each filter one by one and measure the accuracy drop.
Then remove the filter causing the smallest drop. 
This method requires $N_F$ evaluations of the network for removing only one filter, then $N_F-1$ evaluation for the second one and so on.
The problem is, this heuristic cannot find the redundancies of two or more filters.
Imagine that two filters "work" against each other~\cite{doi:10.1080/09540098908915626} and the their absence does not influence the accuracy at all.
However, removing only one of these filters will cause an accuracy drop, since the other filter loses its pair.
Therefore, analyzing the filters independently can cause wrong importance deductions.

%need to examine the accuracy loss as multiple filters are removed
%need to calculate the importance of the filters but they are not independent
%az egyuttes importance has to be calculated
To address these issues, we propose a new filter importance norm where the joint impact of the filters are considered:
We specify a set of filter ensembles -- each filter ensemble is defined by a binary mask vector -- and assign a performance indicator to each.
Assuming linear dependence between the filter ensembles we rank the filters using a linear and additive method and remove the least important ones.

%First we present how to calculate the importance of filters in a single layer.
%Next, we show how to find the optimal pruning percentage in the layer based on the importance values.
%Finally, we describe the process which prunes the entire network.

%TODO compared to soft filter pruning, our approach is better, since it is not required to reach the same pruning level in different layers - some of them could be more important, some of them could be less important - adapts to this

\subsection{Importance of filters in a single layer}\label{sub_sec:filter_importance}
%dataset D and f neural network
Consider $\mathcal{D}_{train} = \{( \pmb{x}_1, y_1), ..., (\pmb{x}_N, y_N)\}$ as a dataset where each $\pmb{x}_i$ denote an input image and $y_i$ its corresponding ground truth label.
Let $f(\pmb{x}|\mathcal{W})$ denote a convolutional neural network.
Here $\pmb{x}$ is the input image and $\mathcal{W} = \{W^1, ..., W^L\}$ denotes the network parameters where the $l-th$ layer contains $N_l$ filters with $W^l = \{w^l_1, ..., w^l_{N_l}\}$ filter matrices.
Suppose that $f(\pmb{x}|\mathcal{W})$ is trained on $\mathcal{D}_{train}$ by minimizing the empirical loss:
\begin{equation}\label{eq:loss}
	\mathcal{L}(f(\cdot | \mathcal{W}), \mathcal{D}_{train}) = \dfrac{1}{N} \sum_{i=1}^{N} C(f(\pmb{x}_i ; \mathcal{W}), y_i)
\end{equation}
where $C(\cdot, \cdot)$ denotes the error function such as cross-entropy loss.

Our goal is to assign values to each $w^l_i$ filter matrix in layer $l$, indicating its importance relative to other filters.
To achieve this, we train a linear model $\mu(\pmb{z}; \pmb{\theta}) = \pmb{\theta}^T \cdot \pmb{z}$ with coefficients $\pmb{\theta} = \{\theta_1, \theta_2, ..., \theta_{N_l}\}$ and use these coefficients as importance values of the filter matrices (e.g. $\theta_i$ as the importance value of filter matrix $w^l_i$).

To train $\mu$, we  create a $\mathcal{D}_\mathrm{mask} = \{( \pmb{z}_1, s_1 ), \ldots (\pmb{z}_M, s_M)\}$ dataset.
Here each $\pmb{z}_i \in \{0, 1\}^{N_l}$ is a binary mask that denotes the filters of the $l-th$ layer which has to be turned off temporarily -- thus each $\pmb{z}_i$ mask specifies a filter ensemble. 
We generate these masks randomly, such that each $z_i$ contains $P \cdot N_l$ zero values.
We set the value of $P$ to 0.3 based on the results of different experiments with varying $P$.
The ``output'' $s_i$ denotes the target value corresponding to mask $\pmb{z}_i$: it is the network score -- the associated performance -- given the filter ensemble specified by $\pmb{z}_i$:
\begin{align}
    \label{eq:score_func}
	s_i =& 1-\dfrac{\mathcal{L}_i - \mathcal{L}_{\min}}{\mathcal{L}_{\max}-\mathcal{L}_{\min}},
    \text{where }\\
    &\mathcal{L}_i = \mathcal{L}(f( \cdot | \pmb{z}_i, \mathcal{W}), \mathcal{D}_\mathrm{train}) \nonumber\\
    &\mathcal{L}_{\min} = \min_{i} \mathcal{L}_i \quad 
    \mathcal{L}_{\max} = \max_{i} \mathcal{L}_i
    \nonumber
\end{align}
The goal of score function in Equation~\eqref{eq:score_func} is to assign high values for those $\pmb{z}_i$ masks where the loss is small (the network performs well) but assign small scores for high losses.
It assigns score 1 to $z_i$ where $\mathcal{L}_i$ is the smallest and score 0 to $z_i$ where $\mathcal{L}_i$ is the largest.
Given $\mathcal{D}_{mask}$, we construct a $\pmb{Z} = [z_1, ... z_M]^T$ matrix where each row of  the matrix is a $\pmb{z_i}$ mask and $\pmb{s} = [s_1, ... s_M]^T$ is a column vector whose elements are the different score values.
We solve the equation $\pmb{Z} \cdot \pmb{\theta} = \pmb{s}$ by computing $\pmb{\theta}$ that minimizes the Euclidean 2-norm:
\begin{equation}\label{eq:loss_mu}
	\mathcal{L}_{\mu} = \lVert \pmb{s} - \pmb{Z} \cdot \pmb{\theta} \rVert ^2
\end{equation}
Finally, we use $\theta_1, ... \theta_{N_l}$ coefficients as importance values of the filters in the neural network.

\subsection{Pruning filters in a single layer}\label{sub_sec:filter_pruning}
Once the filter importances are calculated we can remove the $m$ most unimportant filters from the layer.

% problem with the optimal prunign ratio
The question is how to find the optimal value of $m$?
Removing too much filters from the layer could result in a huge accuracy drop from which we cannot restore the original accuracy.
Removing too few filters will let the retraining process be successful but it can significantly increase the pruning time.

% \alpa hyperparameter
To address this problem we introduce an $\alpha$ threshold which denotes the maximum allowed accuracy drop on a $\mathcal{D}_{val}$ dataset.
We sort the filters based on their $\theta$ importance value and evaluate the network accuracy on $\mathcal{D}_{val}$ dataset as more and more filters  are removed from the layer.
%TODO this is not precise. Basically what we do is sort the filters based on theta, then remove the $p$ least important filter and calculate the accuracy on D_val. If the accuracy is below the threshold we reactive one filter and measure the accuracy again. 
We stop the search as the $\alpha$ threshold is reached.
In our experiments we tune $\alpha$ as a hyperparameter for the best compression results. 

Once the optimal pruning ratio is found, we remove the $m$ least important filters from the layer and finetune the remaining parameters on $\mathcal{D}_{train}$ dataset.

\subsection{Network pruning}
\label{subsec:network_pruning}

In Section~\ref{sub_sec:filter_importance} and ~\ref{sub_sec:filter_pruning} we have shown how we calculate the importance of filters using the linear model and how we find the optimal pruning ratio based on these importance values.
In this section we present our entire pruning process.

%starting from the last layer
As Algorithm~\ref{alg:prune_network} shows, we can start the pruning process either from the first layer ($l = 1$) or from the last layer ($l=L$).
Different pruning directions produce different results.
Since filters in the last layers contains much more parameters than the ones in the first layers (e.g. in ResNet-20 filters in the last layer contain $3 \times 3 \times 64=576$ parameters while filters in the first layers only $3 \times 3\times 16 = 144$), removing one filter from the end of the network results in a bigger drop in the network size.
However, if the goal is to reduce the FLOPs, it is more beneficial to start from the beginning: removing one filter from the first layers reduces the number of FLOPs by $16 \times 3 \times 3 \times 32 \times 32 = 147456$ while this drop in the last layer would be only $64 \times 3 \times 3 \times 8 \times 8 = 36864$.
So removing a filter from the first layers results larger drop in the number of FLOPS compared to filter removals from the last layers.

Choosing either the first or the last layer, we generate a $\mathcal{D}_{mask}$.
Next, we calculate $\pmb{\theta}$ by minimizing Equation~\ref{eq:loss_mu} and find the optimal pruning ratio $m$ presented in Section~\ref{sub_sec:filter_pruning}. 
Next filter pruning is applied based on $\pmb{\theta}$ and $m$ and the pruned network is fine-tuned on $\mathcal{D}_{train}$ dataset to recover the accuracy lost by pruning.
Finally, we move to the next layer and repeat the previous steps until some predefined stopping condition is reached (target compression rate is reached or no more filters can be removed without significant accuracy drop).
Note that layers can be pruned multiple times if stopping condition is not reached.

\begin{algorithm}
	\caption{Network pruning}\label{alg:prune_network}
	\begin{algorithmic}[1]
		\Require $f$ pre-trained network, $\mathcal{D}_{train}$ and $\mathcal{D}_{val}$ datasets 
		\State $l \gets$ index of first or last layer
		\While {stopping condition not met}
			\State generate $\mathcal{D}_{mask}$ for layer $l$
			\State calculate $\pmb{\theta}$ by minimizing equation~\ref{eq:loss_mu}
			\State find optimal $m$ based on $\pmb{\theta}$ and $\mathcal{D}_{val}$
			\State remove $m$ least important filters from layer $l$
			\State fine-tune $f$ on $\mathcal{D}_{train}$
			
			\State $l \gets$ next layer index
		\EndWhile
	\end{algorithmic}	
\end{algorithm}

%whta is the maximum value of accuracy drop?
% the one for which the network can be retrained
% we find this value empirically

%find and remove unimportant and redundant neurons/filters
%after this pruning might be necessary

%\subsection{Algorithm}

\section{Experiments}\label{sec:experiments}
First, we test our pruning algorithm on a small network, trained on the synthetic XOR dataset. 
On this dataset, the minimal network structure which is necessary to solve the problem is known. 
The goal here is to test the pruning algorithm whether it finds this optimal solution or not.
In the next experiment we move to the ResNet architecture trained on the CIFAR-10 dataset.
In this case we compare our pruning results with the results of different state-of-the art methods published in the literature.

\subsection{Simple FCN on a synthetic dataset}
%what is the xor problem
\subsubsection*{XOR dataset}

\begin{figure}[t]
	\centering
	\vspace*{-.5cm}
    \begin{tikzpicture}
	\node[anchor=south west, inner sep=0,outer sep=0,opacity=.7] (image) at (0,0) {\includegraphics[width=\linewidth]{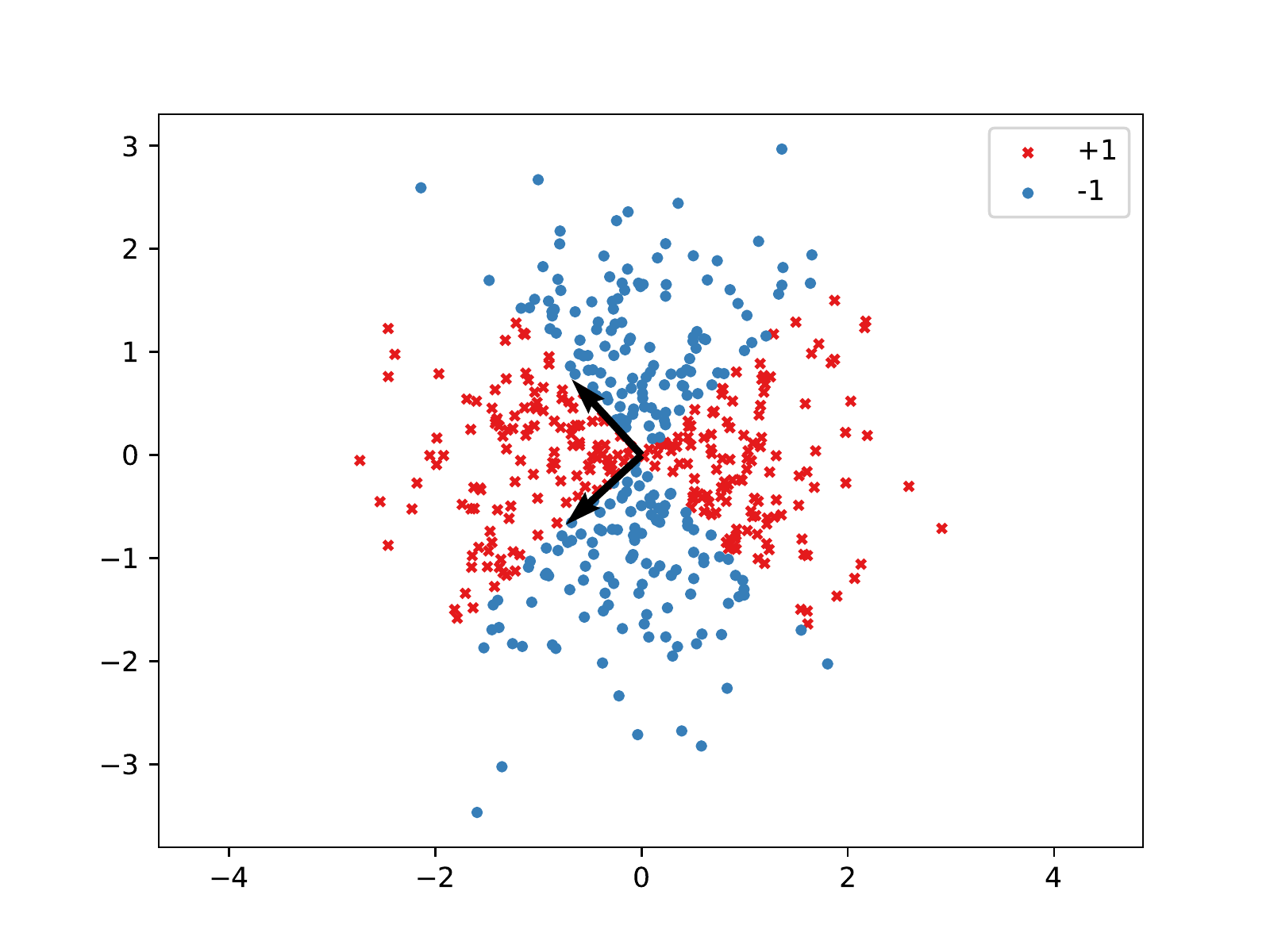}};	
	\begin{scope}[x={(image.south east)}, y={(image.north west)}]
%		\draw[help lines, color=black] (0,0) grid (1,1);
		\node[anchor=south west] at (0.46,0.57) {\LARGE $\pmb{a}$};
		\node[anchor=south west] at (0.46,0.39) {\LARGE $\pmb{b}$};
	\end{scope}
	\end{tikzpicture}
	\vspace*{-1cm}
	\caption{Example for the toy dataset where samples are generated using the orthonormal $\pmb{a}$ and $\pmb{b}$ vectors. }
	\label{fig:xor}
\end{figure}

Our synthetic inputs are two-dimensional and their labels are +1 and -1.
We generate 2 random unit vectors: $\pmb{a}$ and $\pmb{b}$, such that $\pmb{a} \text{ and } \pmb{b}$ are orthonormal.
Next, we generate $\mathcal{D}$ with $N$ pairs of objects $(\pmb{x}^i, y^i)_{i=1}^N$.
Here $\pmb{x}^i$ is sampled from $\mathcal{N}(\pmb{\mu}, \, \pmb{\varSigma})$, where $\pmb{\mu} = [0, 0]^T$,  $\pmb{\varSigma} = [[1, 0], [0, 1]]$ and $y^i$ is the label of $\pmb{x}^i$ such that: 
\begin{equation}
	y^i = \mathrm{sign}\left( (\pmb{a}\cdot \pmb{x}^i)(\pmb{b}\cdot \pmb{x}^i) \right),
\end{equation}
where $\cdot$ denotes the dot product. The above equation projects $\pmb{x}^i$ to $\pmb{a}$ and $\pmb{b}$ and use the sign of their products as the label of the point.
Figure~\ref{fig:xor} shows an example from this XOR dataset.

%include figure

\begin{figure*}[t]
	\centering
	\vspace*{-1cm}
	\begin{tabular}{@{}c@{}c@{}}
	\begin{minipage}{.5\textwidth}
		\includegraphics[width=\textwidth]{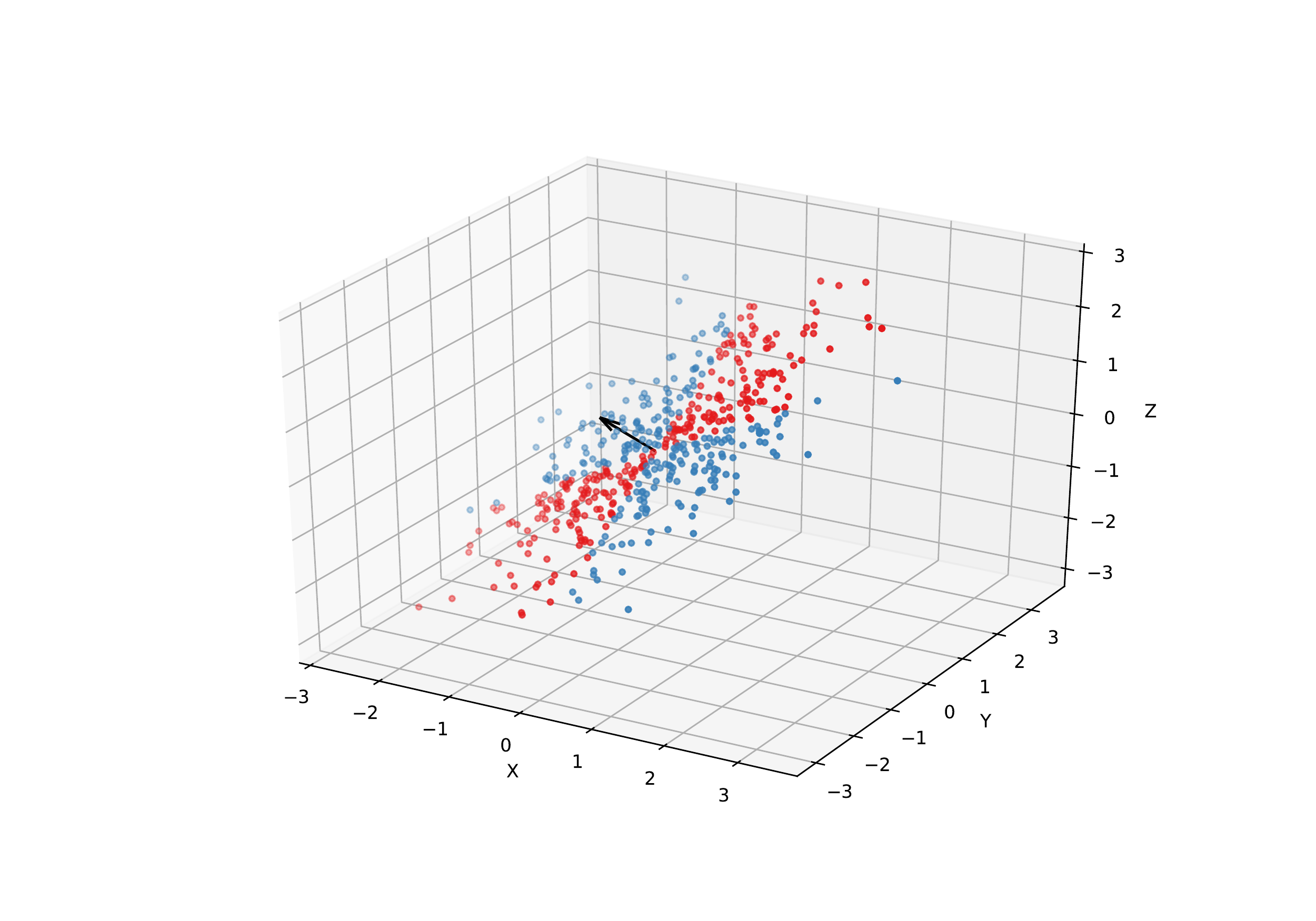}
	\end{minipage}\hfill
	\begin{minipage}{.5\textwidth}
		\includegraphics[width=\textwidth]{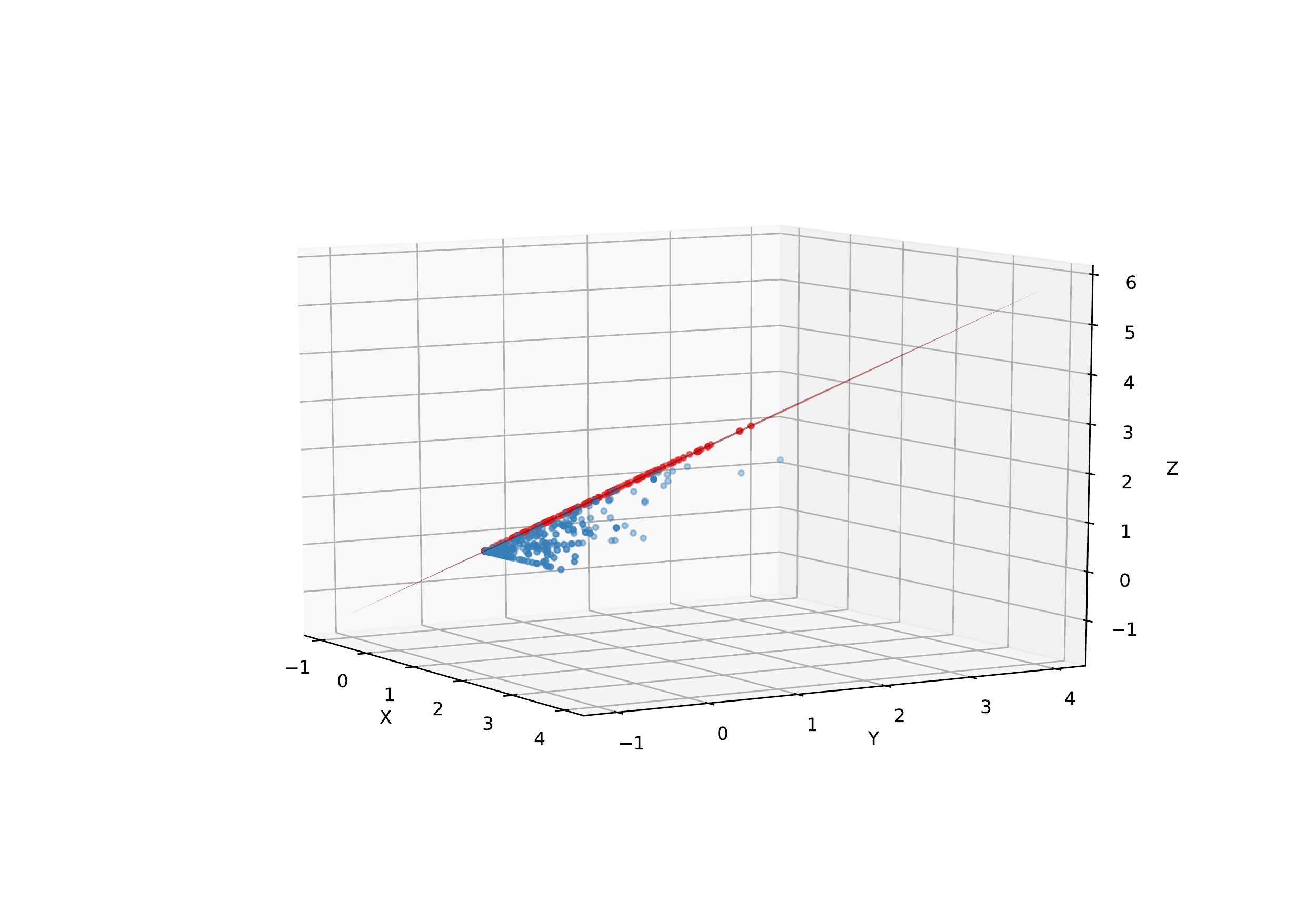}
	\end{minipage}
    \end{tabular}
    \vspace*{-1cm}
	\caption{Illustration of the separation using a deep architecture. (a) Transformed representation of the data and the surface norm; (b)Transformed representation of the data after ReLU activation. This way the data is linearly separable.}
	\label{fig:xor_transformation}
\end{figure*}

%\subsubsection*{Optimal network architecture}
A possible architecture that separates the data is a network with one hidden layer which contains 3 neurons with ReLU activation functions.
In this case, the parameters of the network should be the following.
Let $\pmb{w}^1_i = [w^1_{i,1}, w^1_{i,2}]^T$ denote the parameters of i-th neuron in the hidden layer.
By assigning the values of $\pmb{a}$ and $\pmb{b}$ to $\pmb{w}^1_1$ and $\pmb{w}^1_2$, the neurons rotate the data such that its four regions will be placed in the four quadrant of the XOY plane. %TODO reformulate this
More precisely, the points with positive labels are aligned on the first and third quadrants and points with negative labels are in the second and fourth quadrant.
By assigning $\frac{\pmb{a} + \pmb{b}}{||\pmb{a} + \pmb{b}||}$ to $\pmb{w}^1_3$, the third neuron rotates the data by 45 degree around the X axis.
This new representation is shown in Figure~\ref{fig:xor_transformation}.a.

We denote this transformation as:
\begin{equation}\label{eq:affine_transformation}
	f(\pmb{x}^i, \pmb{W}^1) = \pmb{W}^1\pmb{x}^i  ,
\end{equation}
where $\pmb{W}^1 = [\pmb{w}^1_1, \pmb{w}^1_2, \pmb{w}^1_3]^T$.
This way, the transformed data lies on a plane whose norm is given by:
\begin{align}
\pmb{n}_{plane} = \frac{\partial f}{\partial x_1} \times	\frac{\partial f}{\partial x_2}
= [w^1_{1,1}, w^1_{2,1}, w^1_{3,1}]^T \times [w^1_{1,2}, w^1_{2,2}, w^1_{3,2}]^T,
\end{align}

Adding the ReLU nonlinearity to the hidden layer, the points with positive label will remain on the plane (part of them moved to the origin) while points with negative label will be moved under the plane.
In this case - as Figure~\ref{fig:xor_transformation}.b shows - the data can be separated with a hyperplane.

Given the linearly separable transformed representation, it is easy to calculate the parameters of the hyperplane (the parameters of the output neuron denoted by $\pmb{w}^2_1$), which separates the data.
This hyperplane has to be parallel to the plane defined by $\pmb{n}_{plane}$.
So assigning $\pmb{n}_{plane}$ to $\pmb{w}^2_1$ and setting the bias to a reasonably small value the output neuron will separate the transformed data.

%why is better to use more than 3 neurons in the hidden layer

\subsubsection*{Experimental settings and results}
%pruning results
%To test our pruning method, we train a simple FCN on the above presented dataset and try to decrease its size with the pruning algorithm.
%The used network contains a hidden layer with 10 neurons and an output layer with a single neuron. 
%We use ReLU activation function in the hidden layer and a sigmoid in the output layer.

First, we compare the training success of different networks on the above presented dataset.
We define a training event successful, if it can reach a predefined accuracy value (we set this threshold to 95\% in our experiments) and the training failed, if the accuracy is below this value. 
Sub-optimal solutions at this dataset produce around $50\%$ (the network outputs the same label for all the data) or $75\%$ (three quarters of the data are labeled with one label and the last with the opposite label) accuracy.

\begin{table}
	\centering
	\begin{tabular}{|c | c | c|} 
		\hline
		Network & Method & Percentage of success \\
		\hline
		\multicolumn{3}{|c|}{Training}\\
		\hline
		FCN3 & - & 40.4\%  \\ 
		FCN10 & - & 99.5\% \\
		\hline
		\multicolumn{3}{|c|}{Prunnig} \\
		\hline
		FCN10 & random & 39.8\\
		FCN10 & ours - one-shot & 82.6\%\\ %p = 0.9
		FCN10 & ours - iterative & 88\%\\
		
		\hline
	\end{tabular}
	\caption{Percentage of success in different training and pruning scenarios. FCN3 is a fully connected network with 3 neuron in its hidden layer while FCN10 is a similar one with 10 neurons.}
	\label{tabel:xor_results}
\end{table}

We compare the training success of two different networks: 
FCN3 contains a hidden layer with 3 neurons - just as many parameters as it is required to separate the data; FCN10 contains 10 neurons in its hidden layer. 
In both networks ReLU is used in the hidden layer and sigmoid at the output neuron.
With each network, we run 1000 experiments and at each experiment we randomly reinitialize the network and generate a new dataset.
\begin{algorithm}[b]
	\caption{}\label{alg:xor_pruning}
	\begin{algorithmic}[1]
		\Require $nr\_experiments, pruning\_type$
		\For {$i = 1 \text{ to } nr\_experiments$}
		\State initialize random $\mathcal{D}_{XOR}$ and FCN10
		\State train FCN10 on $\mathcal{D}_{XOR}$
		\If {$pruning\_type$ == "one-shot"}
		\State prune 7 neurons from FCN10
		\State retrain pruned FCN10
		\ElsIf {$pruning\_type$ == "iterative"}
		\For {i in [3. 2, 2]}
		\State prune $i$ neurons from FCN10
		\State retrain the pruned FCN10
		\EndFor
		\EndIf
		\If {$accuracy > 95\%$}
		\State $nr\_success \gets nr\_success + 1$
		\EndIf
		\EndFor
		
		\State $\text success\_rate \gets \frac{nr\_succes}{nr\_experiments}$	
	\end{algorithmic}
\end{algorithm}
The first 2 lines of Table~\ref{tabel:xor_results} shows the training success of the different networks.
In case of FCN3 the success rate is only $40.4\%$.
This means that in around $60\%$ of the experiments the training gets stuck in some local optimum and the network cannot separate the data.
However, the network with 10 neurons can learn the problem with a success rate of $99.5\%$ percent.

% This is probably not necessary
%This can be due to different reasons.
%One explanation is related to the lottery thicket hypothesis, presented in~\cite{DBLP:journals/corr/abs-1803-03635}.
%Based on this,  a network with many parameters has higher chance to have "winning tickets" - subnetworks with lucky initialization with which the network can be trained on the problem.
%Another explanation is that adding more parameters to the network convexifies the loss surface - cite
%As Figure~\ref{fig:NN3_loss_surface} shows, a neural network with few parameters (3 neuron in the hidden layer) has a loss surface with multiple local/global minimas.
%However, a network with more parameters (Figure~\ref{fig:NN10_loss_surface} - 10 neuron in the hidden layer) has a loss surface which is closer to the required convex shape.

%\begin{figure*}[htp]
%	\centering
%	\begin{subfigure}{0.45\textwidth}
%		\includegraphics[width=\textwidth]{figures/NN3_loss_surface}
%		\caption{Loss surface of neural network with 3 neurons in the hidden layer}
%		\label{fig:NN3_loss_surface}
%	\end{subfigure}\hfill
%	\begin{subfigure}{0.45\textwidth}
%		\includegraphics[width=\textwidth]{figures/NN10_loss_surface}
%		\caption{Loss surface of neural network with 10 neurons in the hidden layer}
%		\label{fig:NN10_loss_surface}
%	\end{subfigure}
%\end{figure*}

\begin{figure*}[t]
	\centering
	\begin{minipage}{.33\textwidth}
		\begin{tikzpicture}[node distance=0.7cm]
		\node (start) [invisible] {};
		\node (conv1) [layer, below of=start, yshift=-0.3cm] {conv 3x3};
		\node (bn1) [layer, below of=conv1] {batch norm};
		\node (relu1) [layer, below of=bn1] {relu};
		\node (conv2) [layer, below of=relu1, yshift=-0.6cm] {conv 3x3};
		\node (bn2) [layer, below of=conv2] {batch norm};
		\node (plus) [concat, below of=bn2] {+};
		\node (relu2) [layer, below of=plus] {relu};
		\node (end) [invisible, below of=relu2] {};
		\draw [arrow] (start) -- node [anchor=east] {$x$} (conv1);
		\draw [arrow] (conv1) -- (bn1);
		\draw [arrow] (bn1) -- (relu1);
		\draw [arrow] (relu1) -- node [anchor=west, xshift=1.6cm] {$x$ identity} node [anchor=east, xshift=-0.9cm] {$\mathcal{F}(x)$}(conv2);
		\draw [arrow] (conv2) -- (bn2);
		\draw [arrow] (bn2) -- (plus);
		\draw [arrow] (plus) -- node[anchor=south east, xshift=-0.2cm] {$\mathcal{F}(x) + x$} (relu2);
		\draw [arrow] (relu2) -- (end);
		\draw [arrow] (start) |- ([yshift=0.5cm, xshift=0.5cm]conv1.north east) |- (plus);
		\end{tikzpicture}
	\end{minipage}
	\begin{minipage}{.33\textwidth}
		\begin{tikzpicture}[node distance=0.7cm]
		\node (start) [invisible] {};
		\node (conv1) [layer, below of=start, yshift=-0.3cm] {conv 3x3};
		\node (bn1) [layer, below of=conv1] {batch norm};
		\node (relu1) [layer, below of=bn1] {relu};
		\node (mask1) [red_layer, below of=relu1] {mask};
		\node (conv2) [layer, below of=mask1, yshift=-0.6cm] {conv 3x3};
		\node (bn2) [layer, below of=conv2] {batch norm};
		\node (plus) [concat, below of=bn2] {+};
		\node (relu2) [layer, below of=plus] {relu};
		\node (mask2) [red_layer, below of=relu2] {mask};
		\node (end) [invisible, below of=mask2] {};
		\draw [arrow] (start) -- node [anchor=east] {$x$} (conv1);
		\draw [arrow] (conv1) -- (bn1);
		\draw [arrow] (bn1) -- (relu1);
		\draw [arrow] (relu1) -- (mask1);
		\draw [arrow] (mask1) -- node [anchor=west, xshift=1.6cm] {$x$ identity} node [anchor=east, xshift=-0.9cm] {$\mathcal{F'}(x)$}(conv2);
		\draw [arrow] (conv2) -- (bn2);
		\draw [arrow] (bn2) -- (plus);
		\draw [arrow] (plus) -- node[anchor=south east, xshift=-0.2cm] {$\mathcal{F'}(x) + x$} (relu2);
		\draw [arrow] (relu2) -- (mask2);
		\draw [arrow] (mask2) -- (end);
		\draw [arrow] (start) |- ([yshift=0.5cm, xshift=0.5cm]conv1.north east) |- (plus);
		\end{tikzpicture}
	\end{minipage}	
	\begin{minipage}{.33\textwidth}
		\begin{tikzpicture}[node distance=0.7cm]
		\node (start) [invisible] {};
		\node (conv1) [layer, below of=start, yshift=-0.3cm] {conv 3x3};
		\node (bn1) [layer, below of=conv1] {batch norm};
		\node (relu1) [layer, below of=bn1] {relu};
		\node (mask1) [red_layer, below of=relu1] {mask};
		\node (conv2) [layer, below of=mask1, yshift=-0.6cm] {conv 3x3};
		\node (bn2) [layer, below of=conv2] {batch norm};
		\node (mask2) [red_layer, below of=bn2] {mask};
		\node (plus) [concat, below of=mask2] {+};
		\node (relu2) [layer, below of=plus] {relu};
		\node (end) [invisible, below of=relu2] {};
		\draw [arrow] (start) -- node [anchor=east] {$x$} (conv1);
		\draw [arrow] (conv1) -- (bn1);
		\draw [arrow] (bn1) -- (relu1);
		\draw [arrow] (relu1) -- (mask1);
		\draw [arrow] (mask1) -- node [anchor=west, xshift=1.6cm] {$x$ identity} node [anchor=east, xshift=-0.9cm] {$\mathcal{F''}(x)$}(conv2);
		\draw [arrow] (conv2) -- (bn2);
		\draw [arrow] (bn2) -- (mask2);
		\draw [arrow] (mask2) -- (plus);
		\draw [arrow] (plus) -- node[anchor=south east, xshift=-0.2cm] {$\mathcal{F''}(x) + x$} (relu2);
		\draw [arrow] (relu2) -- (end);
		\draw [arrow] (start) |- ([yshift=0.5cm, xshift=0.5cm]conv1.north east) |- (plus);
		\end{tikzpicture}
	\end{minipage}
	%\vspace*{-1cm}
	\caption{Residual blocks in ResNet architecture. Left: original residual block. Middle: residual block with mask layer after the shortcut connection. Right: residual block with mask layer before the shortcut connection.}
	\label{fig:residual_block}
\end{figure*}
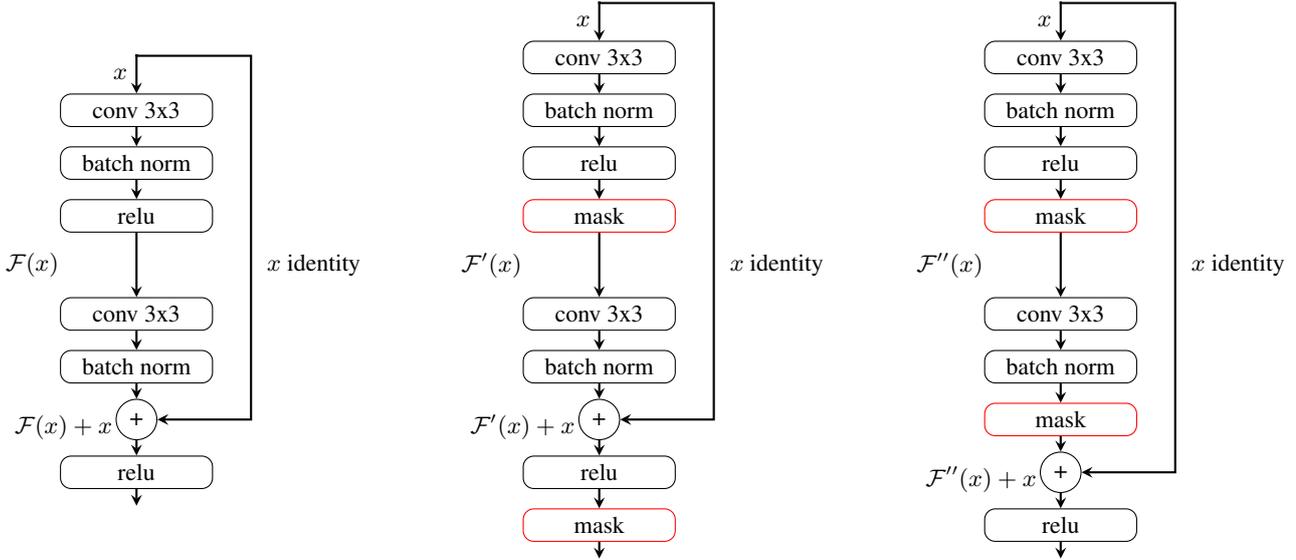

Next, we apply our pruning algorithm on the trained FCN10 network.
Here, the goal of the experiment is to measure the success rate of the pruning process: from a set of experiments how many times can the method find the optimal network structure which contains only 3 neurons but still can separate the data?
Here we evaluate two different versions of our pruning algorithm. 
%TODO TODO
The first is one-shot pruning, which simultaneously removes all the seven neurons then retrains the network. The second is iterative pruning which progressively removes more and more neurons and applies retraining between the pruning steps. 
In this case we iteratively reduce the number of neurons to 7, 5 and finally to 3.

Similarly to the previous case, we repeat each pruning process 1000 times and calculate the success rate of the pruning. 
We define a pruning process successful if the pruned and retrained network reaches at least $95\%$ accuracy on the XOR dataset.

The second part of Table~\ref{tabel:xor_results} shows the results of this experiment. 
As a reference, we also measured the success rate of random pruning (with retraining).
As the table shows, the random pruning achieves $39.8\%$ success rate on the \textit{XOR} dataset while our method achieves $82.6\%$ in case of one-shot pruning.
Moreover, in case of iterative pruning this value reaches $88\%$.

%
% 3 neurons
% 10 neurons
% one shot pruning
% iterative pruning - 
% 	- no reinit
% 	- retrain
%	- retrain and reinit
%

\subsection{ResNet architectures on the CIFAR-10 data}

%possible extension:
%comapre pruned and retrained network with the same pruned network architecture but the network is trained from scratch. Like in Pruning neural networks : is it time to nip it in the bud ?

% resnet in general - skip connections, why is that good
% how we apply pruning in resnet?

% results
% % results compared with other papers
% % pruned filters per layer - table
% % forward / backward pruning
% % prining when the mask is after the skip connection?

Beside the synthetic toy data, we also evaluate our pruning method on the ResNet~\cite{DBLP:conf/cvpr/HeZRS16} architectures trained on CIFAR-10~\cite{cifar10} dataset. 
Here we analyze the forward and backward pruning directions, as well as the pruning possibilities of the residual block.

\subsubsection*{Dataset and experimental settings}

CIFAR-10 is a dataset which contains 60000 $32 \times 32$ color images, categorized into 10 classes.
These images are split into train and test sets such that the first has 50000 while the second has 10000 samples.
In our experiment, we further split the first set to get a train and validation set with 45000 and 5000 images.

ResNet~\cite{DBLP:conf/cvpr/HeZRS16} is a convolutional neural network architecture which uses residual blocks and "shortcut connections" for better propagation of the error signal.
Compared to other convolutional neural networks (such as VGG~\cite{Simonyan15}), ResNet has a more compact architecture. This model was developed for the ImageNet~\cite{imagenet_cvpr09} dataset but it has a smaller version for CIFAR-10 as well. 
The building block of each ResNet architecture is the residual block, which contains a series of stacked convolutional, nonlinear and batch normalization layers.
However, its goal is not to learn the target distribution $\mathcal{H}(x)$, but to learn the residual function $\mathcal{F}(x) = \mathcal{H}(x) - x$, where $x$ is the input feature map of the residual block. 
This is accomplished by applying a "shortcut connection" - performed by identity mapping - between the input and output of the residual block.
Combining the output of the shortcut connection and the output of the residual block the target distribution is approximated in the form of $\mathcal{H}(x) = \mathcal{F}(x) + x$.

In ResNet, each residual block contains two sets of convolutional, batch normalization and ReLU layers, such that the output of a layer is fed into the input of the next layer.
However,  the output of the shortcut connection is also added to the output produced by the second batch norm layer. %TODO TODO
This is shown on the left side of Figure~\ref{fig:residual_block}.
 
\begin{table*}[t]
	\centering
	% Preview source code for paragraph 0
	
	\begin{tabular}{|c|c|c|c c c c|c c c c|}
		\hline 
		Layer & \#Filters & \#FLOPs & \multicolumn{4}{c|}{\#Filters $\downarrow$} & \multicolumn{4}{c|}{Params./FLOPs $\downarrow$ (\%)} \tabularnewline 
		&    &       & BF & BB & AF & AB & BF    &BB     & AF    & AB \tabularnewline
		\hline
		0  & 16 & 0.44M & 1  & 0  & 2  & 0  & 6.25  & 0.00  & \textbf{12.50} & 0.00 \tabularnewline
		\hline 
		1  & 16 & 2.35M & 6  & 1  & 6  & \textbf{9}  & 41.41 & 6.25  & 45.31 & \textbf{56.25} \tabularnewline  
		2  & 16 & 2.35M & \textbf{4}  & 0  & 3  & 0  & 53.12 & 6.25  & 49.22 & \textbf{56.25} \tabularnewline
		3  & 16 & 2.35M & 7  & 1  & 6  & \textbf{8}  & 43.75 & 6.25  & 49.22 & \textbf{50.00} \tabularnewline
		4  & 16 & 2.35M & \textbf{6}  & 0  & 1  & 0  & \textbf{64.84} & 6.25  & 41.41 & 50.00 \tabularnewline
		5  & 16 & 2.35M & 0  & 0  & \textbf{1}  & 0  & 0.00  & 0.00  & \textbf{12.11} & 0.00  \tabularnewline
		6  & 16 & 2.35M & 0  & 0  & 0  & 0  & 0.00  & 0.00  & \textbf{6.25}  & 0.00  \tabularnewline
		\hline 
		7  & 32 & 1.17M & 5  & 0  & \textbf{6}  & 2  & 15.62 & 0.00  & \textbf{18.75} & 6.25  \tabularnewline 
		8  & 32 & 2.35M & 1  & 0  & \textbf{4}  & 0  & 18.26 & 0.00  & \textbf{28.91} & 6.25  \tabularnewline
		9  & 32 & 2.35M & \textbf{14} & 7  & 6  & 13 & \textbf{43.75} & 21.88 & 28.91 & 40.62 \tabularnewline
		10 & 32 & 2.35M & \textbf{20} & 9  & 0  & 0  & \textbf{78.91} & 43.85 & 18.75 & 40.62 \tabularnewline
		11 & 32 & 2.35M & 0  & 1  & \textbf{20} & 0  & 0.00  & 3.12  & \textbf{62.50} & 0.00  \tabularnewline
		12 & 32 & 2.35M & 2  & \textbf{6}  & 0  & 0  & 6.25  & 21.29 & \textbf{62.50} & 0.00  \tabularnewline
		\hline 
		13 & 64 & 1.17M & 0  & 16 & 9  & \textbf{19} & 0.00  & 25.00 & 14.06 & \textbf{29.69} \tabularnewline
		14 & 64 & 2.35M & 4  & 6  & \textbf{7}  & 0  & 6.25  & \textbf{32.03} & 23.46 & 29.69 \tabularnewline
		15 & 64 & 2.35M & \textbf{34} & 16 & 26 & 18 & \textbf{53.12} & 25.00 & 47.12 & 28.12 \tabularnewline
		16 & 64 & 2.35M & \textbf{30} & 23 & 5  & 1  & \textbf{75.10} & 51.95 & 45.26 & 29.25 \tabularnewline
		17 & 64 & 2.35M & 0  & \textbf{21} & 0  & 16 & 0.00  & \textbf{32.81} & 7.81  & 26.17 \tabularnewline
		18 & 64 & 2.35M & 24 & 23 & 21 & \textbf{27} & 37.50 & \textbf{56.96} & 32.81 & 56.64 \tabularnewline
		\hline  
		Total & 688 & 40.5M & 187 & 130 & 123 & 113 & 30.70 / 30.91 & 32.33 / 18.99 & 31.55 / 33.76 & 30.41 / 28.38 \tabularnewline
		\hline
	\end{tabular}
	
	\caption{Percentage of pruned parameters and pruned FLOPs per layer in ResNet-20 by varying the pruning direction and second mask layer position in the residual blocks. Column "\#Filters $\downarrow$" and  "Params./FLOPs $\downarrow$(\%)" shows the number of removed filters and the percentage of the removed parameters/FLOPs in the respective layer in case of the four experiment: mask before the shortcut connection, forward ("BF") and backward ("BB") pruning; mask after the shortcut connection, forward ("AF") and backward ("AB") pruning.}
	\label{table:resnet20}
\end{table*} 

\textbf{Training details:} 
We experiment with the following ResNet architectures: ResNet-20, 32, 56 and 110.
We use the same implementation details and parameters as presented in \cite{DBLP:conf/cvpr/HeZRS16} except from the followings: we insert mask layers into the residual blocks as presented in the next "Pruning details" paragraph; we divide the initial $0.1$ learning rate by 10 at 100, and 150 epochs and stop the training at 200 epochs. 
During training we apply cropping and horizontal flip as data augmentation.
The network is trained on a 45K train set, a 5K validation set is retained for validation purposes during pruning.

\textbf{Pruning details:}
To calculate filter importance values, we insert mask layers into the residual blocks. 
Each mask layer contains a binary vector that zeros out feature maps produced by the convolutional layers.

%Mask layer position
We use two mask layers in each residual block. 
The first one is inserted between the first ReLU and the second convolutional layer. 
This ensures that some of the filters from the first convolutional layer have no effect on the network output.
In case of the second mask layer, we have experimented with two different positions.:
\begin{enumerate*}[label={(\alph*)}, font={\bfseries}]
	\item 
	We put the mask layer after the second ReLU layer (Figure~\ref{fig:residual_block}, middle). 
	This way, some of the filters from the second convolution have no effect on the network output.
	This position also implicates that some of the channels can be removed from the shortcut connection, since they are zeroed out by the mask layer.
	The benefit of this is that some parameters from the next convolutional layer can also be removed. thus the compression ratio will be better.
	However, the experiments show that removing channels from the shortcut connection produces large accuracy drop which is not desired during pruning. 
	
	\item
	The second option is to place the mask layer before the shortcut connection (Figure~\ref{fig:residual_block}, right).
	This means that some feature maps from the convolutional layer are masked but the channels from the skip connections are retained.
	This results in less accuracy drop but less compression ratio as well (the number of output channels of the residual block does not change after pruning).
	
\end{enumerate*}

% pruning parameters: number of masks, pruning probability, accuracy drop thrs, retrainnig epochs
During pruning we follow the process as described in Section~\ref{subsec:network_pruning}.
Starting either from the beginning or from the end of the network, filter importance calculation and pruning is done layer by layer.
To calculate filter importance values, $\mathcal{D}_{mask}$ is created, where the size of the dataset is $10 \times N_l$. 
Each mask in the dataset is generated randomly, such that $p$ percentage of its value is set to 0.
This way, each mask turns off the same number of filters in the layer.
We set the $p$ value to 0.3 in our experiments.
Furthermore to not loose too much accuracy during training, we set the maximum accuracy drop threshold to 0.5\%.
After a layer is pruned, we apply fine-tuning for 10 epochs.
Finally, when no more filters can be removed, we retrain the network for 80 epochs by setting the learning rate to $0.01$ and multiply it by 0.1 in every 20 epochs.

\subsubsection*{Results}

\textbf{Comparison of different pruning methods on ResNet-20: }First, we analyze the effect of our pruning method layer-by-layer on the ResNet-20 architecture.
More specifically, we compare the forward and backward pruning when the mask layer is applied before and after the shortcut connections. 
In case of all four experiments, we remove roughly $30\%$ of the parameters.
Results are shown in Table~\ref{table:resnet20}.

Each row in the table corresponds to one convolutional layer from the ResNet-20 architecture (we do not prune the output layer).
Column "\#Filters" and "\#FLOPs" shows the number of filters and floating point operations in the respective layer. 
"\#Filters $\downarrow$" shows the number of pruned filters in case of the four pruning experiment: "BF" - mask before shortcut connections, forward pruning; "BB" - mask before shortcut connection, backward pruning; "AF" - mask after shortcut connection, forward pruning"; "AB" - mask after shortcut connection, backward pruning.
Finally, column "Params./FLOPs $\downarrow$ (\%)" contains the percentage of pruned parameters and floating point operations. 
Here the two values are equal since removing one filter means the same percentage of parameter and flops reduction.
The last line contains the cumulative values of the corresponding columns. 
%TODO last line param and accuracy drop are not equal - explain why

%filters in the first layers are more important than filters in the last layers
As Table~\ref{table:resnet20} shows, most of the filters are removed from the last layers, which means filters in the first layers are more important.
For example in case of the "BF" experiment 23 filters are removed from the first three residual blocks ($1.74\%$ of the parameters) while 92 ($23.5\%$) are pruned from the last three blocks.
For the "BB" experiment these values are 2 filters against 105 ($0.21\%$ against $29.02\%$).

%importance of pruning direction
\begin{table*}[t]
	\centering
	%		\scriptsize
	\begin{tabular}{|c|ccccc|}
		\hline 
		ResNet 					& Method 				& Acc. before(\%) & Acc. After (\%) & FLOPs $\downarrow$ (\%)& Params. $\downarrow$(\%)\tabularnewline
		 
		\hline 
		\multirow{3}{*}{20} 	& \cite{ijcai2018-309} 	& 92.20 & 90.38 & 42.20 & 41.54 \tabularnewline
		& \cite{he2019filter} 	& 92.20 & 91.99 		& \textbf{54.00}& \textbf{53.59}	\tabularnewline
		& Ours 					& 92.13 & 91.42			& 45.05 		& 46.42\tabularnewline
		 
		\hline 
		\multirow{3}{*}{32} 	& \cite{ijcai2018-309} 	& 92.63 & 92.08 		& 41.5 			& 41.24\tabularnewline
		& \cite{he2019filter} 	& 92.63 & 92.82 		& \textbf{53.2} & \textbf{53.2}\tabularnewline
		& Ours 					& 92.97 & 92.42 		& 46.4	 		& 49.35\tabularnewline
		 
		\hline 
		\multirow{4}{*}{56} 	& \cite{DBLP:journals/corr/LiKDSG16}& 93.04 & 93.06 		& 27.6			& 13.7\tabularnewline
		& \cite{ijcai2018-309} 	& 93.59 & 93.35 		& 47.14 		& 52.6\tabularnewline
		& \cite{he2019filter} 	& 93.59 & 93.49			& 47.14 		& 52.6\tabularnewline
		& Ours 					& 93.44 & 93.18 		& \textbf{57.64}& \textbf{68.14}\tabularnewline
		 
		\hline 
		\multirow{4}{*}{110} 	& \cite{DBLP:journals/corr/LiKDSG16}& 93.53 & 93.3 			& 38.6 			& 32.40\tabularnewline
		& \cite{ijcai2018-309} 	& 93.68 & 93.86 		& 40.8 			& 40.72 \tabularnewline
		& \cite{he2019filter} 	& 93.68 & 93.85			& 52.3	 		& 52.7 \tabularnewline
		& Ours 					& 94.05 & 93.48 		& \textbf{63.68}& \textbf{60.08} \tabularnewline
		\hline 
	\end{tabular}
	\caption{Comparison of pruned ResNet with the results of \cite{DBLP:journals/corr/LiKDSG16, ijcai2018-309, he2019filter}. Columns "Acc. before (\%)" and "Acc. after (\%)" shows the network accuracy before and after pruning while columns "FLOPs $\downarrow$"  and "Params. $\downarrow$" contains the percentage of parameter and floating point operation reduction by the different pruning methods.}
	\label{table:resnets-results}	
\end{table*}

Another important factor here is the pruning direction. 
Starting pruning from the beginning or from the end of the network has its own benefits and drawbacks.
On the one hand, filters in the last layers contain much more parameters than filters in the first layers.
This is because the input tensor of the last layers have more channels (64 for layers 14-18 while layers 1-6 have only 16).
More input channels require more filter channels which increases the number of parameters significantly. 
For example, a filter in the 18th layer contains $64 \times 3 \times 3 = 576$ parameters, while a filter in the 1st layer has only $16 \times 3 \times 3=144$ parameters.
Removing filters from the last layers results higher drop in the network size, therefore the pruning process can reach faster the desired compression level.
Another benefit of the backward pruning is related to filter importance.
Since filters at the beginning are more important, removing them will cause more drop on the accuracy.
This could prevent further pruning from the latter layers where the majority of the parameters are located.
By starting the pruning process from the last layer, one can remove many of the parameters without reducing too much the the accuracy.

On the other hand, a filter from the first layers responsible for more floating point operations than a filter from the last layers.
For example, a filter from layers 1-6 requires $32 \times 32 \times 16 \times 3\times 3 = 147456$ FLOPs while a filter from layers 14-18 only $8 \times 8 \times 64 \times 3 \times 3 = 36864$.
Therefore, removing one filter from the first layers produces more decrease in FLOPs than pruning filters from the last layers.
As the table shows, in case of forward pruning we managed to remove more than $30\%$ of FLOPs while in case of backward pruning this values were only $18.99\%$ and $28.28\%$.

%mask after relu - smaller compression ratio could be achieved
Finally, we also analyze the effect of the mask layer's different position.
As mentioned above, by applying the mask layer after the ReLU operation, one could remove not just the filters from the second convolution but also the appropriate channels from the shortcut connection.
This way, the residual block would produce less feature maps which induce less parameters in the next filters.
Unfortunately, the empirical results show that it is not possible to remove channels from the skip connection without significant accuracy drop.
By analyzing the columns "AF" and "AB", we can see that in most cases no filters (or only a few) are pruned at layers with even indices (these are the second convolutional layers in the residual blocks).
The only exception here is the last layer, since more than 20 filters are pruned in both -- the forward and the backward -- cases.
This is possible since there is no shortcut connection from the output of this layer.

%based on these we prune filters starting from the first layer and place the mask layer before the shortcut connection
Based on the results presented above, we decide to apply the "BF" pruning version in the next experiments.
We choose the forward pruning method since our goal is not only to reduce the network size but also to decrease the floating point operations during inference.
We find more advantageous to insert the second mask layer before the shortcut connection since this way we managed to remove more parameters from the network.
Applying the mask layer after the shortcut connection prevents pruning almost half of the convolutional layers which leads to less compression ratio.

\textbf{Comparison with state-of-the-art methods:}
We test our pruning method on different versions of ResNet and compare the results with different state-of-the-art filter pruning methods.
As Table~\ref{table:resnets-results} shows, our method has comparable results with other filter pruning methods.
In case of the shallower networks -- ResNet-20 and ResNet-32 -- we manage to remove more parameters and FLOPs than \cite{ijcai2018-309} still having higher accuracy.
However, our results are below the results of ~\cite{he2019filter} in terms of FLOPs and parameters reduction as well as in terms of accuracy.
In case of the deeper networks -- ResNet-56 and ResNet-110 -- our method produced smaller networks with less floating point operations.
In case of ResNet-110, we removed $60.08\%$ of the parameters which means $63.68\%$ FLOPs reduction.
This means $7.38\%$ less parameter and $11.38\%$ less floating point operation compared to the results of \cite{he2019filter}.
Furthermore, we managed to remove $68.14\%$ of the parameters in ResNet-56 ($15.54\%$ more than \cite{he2019filter}) and $57.54\%$ of the floating point operations ($10.5\%$ more removed FLOPs compared to ~\cite{he2019filter}).
 
These results validate the effectiveness of our method which can compress significantly the ResNet architectures and produce comparable results with the state-of-the-art methods.

\section{Conclusion and future work}

%new criteria for 
In this paper, we have presented a filter pruning method, which removes filters to accelerate deep neural networks.
The pruning is achieved by a novel filter importance norm that is estimated by considering the change in the network loss as random filters are masked out.
Moreover, the method adaptively determines the number of removable filters per layer, such that the accuracy drop remains below a predefined value.
To show the effectiveness of the filter importance norm, we have evaluated our method on a small fully connected network trained on a small synthetic dataset, as well as on the CIFAR-10 version of the ResNet architecture where we reached comparable results with current state-of-the-art methods.
In the future, we plan to test our pruning technique with networks trained on large-scale data-sets (e.g. ImageNet), as well as combine with other methods such as network quantization.

%\ack We would like to thank the referees for their comments, which
%helped improve this paper considerably

\bibliography{bibliography}
\end{document}